\documentclass[11pt,a4paper]{article}
\usepackage{xgen-report}
\usepackage{placeins,needspace}
\newcommand{\ReportTitle}{Generative World Simulation with DAO and JING}
\newcommand{\ReportAuthors}{\textbf{XGEN Team}}
\newcommand{\ReportDate}{}

\title{\ReportTitle}
\author{\ReportAuthors}
\date{\ReportDate}
\hypersetup{pdftitle={XGEN},pdfauthor={XGEN Team},pdfsubject={Generative World Simulation with DAO and JING}}
\begin{document}
\maketitle
\vspace{-10pt}
\begin{abstract}
We present a Generative World Simulation framework that couples persistent shared world state with continuous first-person audiovisual generation. The framework integrates DAO, a computable world engine, and JING, an Egocentric Interactive Experience Model, to support sustained interaction for humans and agents. DAO maintains the states of characters, objects, and spaces and their relationships, advancing the world according to interaction rules and world time while preserving the consequences of past actions. From this evolving state, DAO derives observer-specific conditions for JING, which combines these conditions with character, object, and scene references, observation history, and current actions to generate each observer’s audiovisual experience. Users can explore environments, manipulate objects, and communicate with characters through viewpoint control and text instructions, with their actions shaping both the shared state and subsequent experiences. By connecting world-state evolution with experience generation in a continuous interaction loop, the framework offers an approach to building interactive worlds that retain their history, evolve through action, and persist beyond any individual observer.
\par\smallskip
{\footnotesize
BLOG: \url{https://xgenlabs.ai/research/generative-world-simulation}\\[2pt]
MODEL: \url{https://huggingface.co/XGENlabs/XGEN-JING}\par}
\end{abstract}

% Keep the teaser directly below the abstract instead of allowing it to float.
\par\vspace{-5pt}\noindent
\begin{minipage}{\linewidth}
\centering
\includegraphics[width=\linewidth]{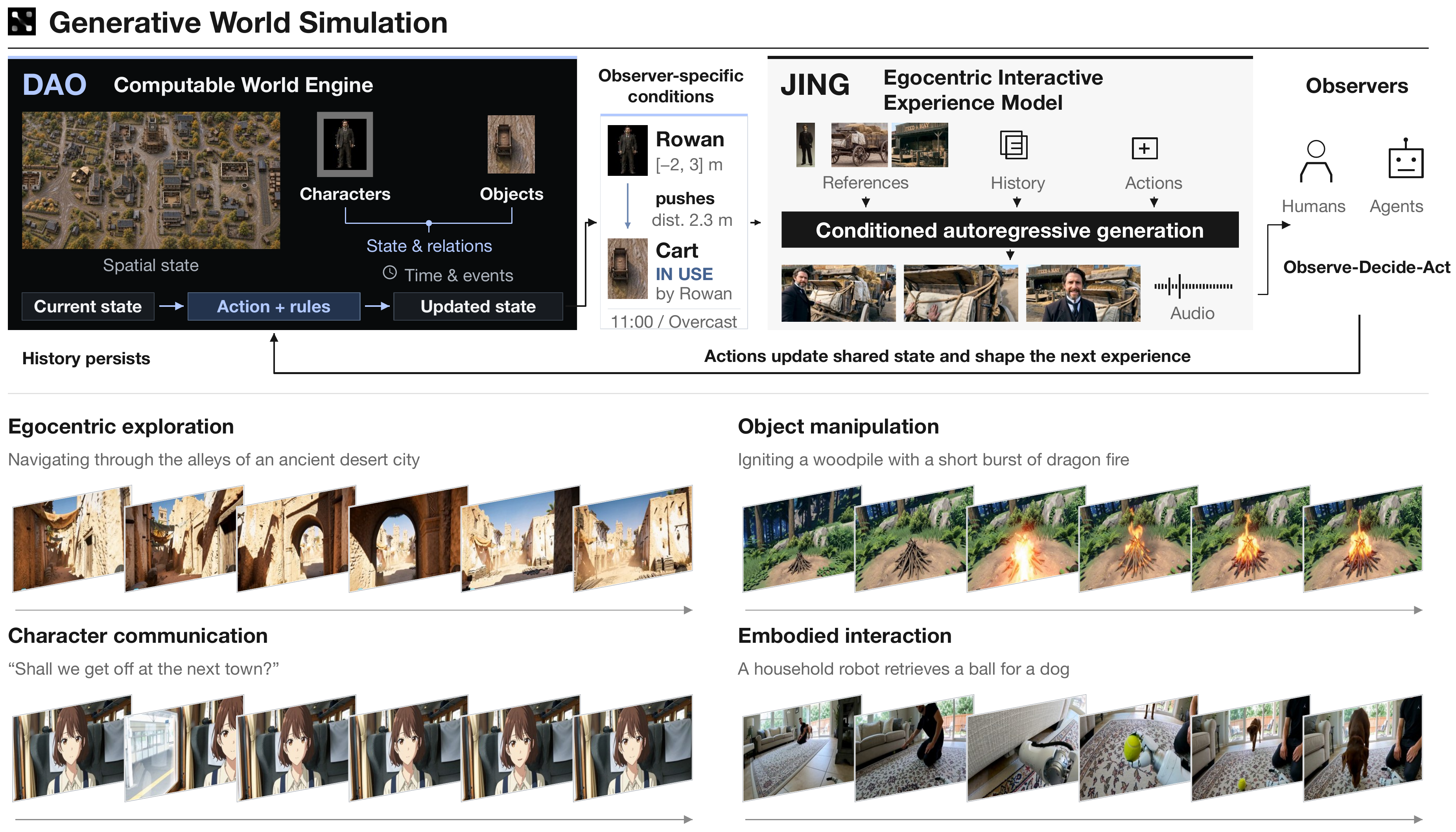}
\captionof{figure}{\textbf{Generative World Simulation.} DAO maintains shared world state and updates it in response to actions; JING generates each observer's first-person audiovisual experience from the resulting conditions. Representative JING trajectories illustrate exploration, object manipulation, character communication, and embodied interaction.}
\label{fig:teaser}
\end{minipage}

% Former layout-only front-page placeholders, retained for reference.
\iffalse
\section*{Research Overview}
\begin{description}[style=nextline,leftmargin=0pt,labelsep=0pt,font=\sffamily\bfseries,itemsep=7pt]
  \item[Problem] Define the research question, the intended setting, and the limitations of existing approaches.
  \item[Approach] Explain the central design choice and how the system's components address the problem.
  \item[Evidence] Connect each contribution to a reproducible comparison, ablation, or analysis.
\end{description}
\vfill
{\small\color{XGENGray}\emph{Layout preview. Technical content and author details remain to be finalized.}\par}
\fi

\clearpage
\tableofcontents
\clearpage
\section{Introduction}
\label{sec:intro}

Video generation models~\cite{wan2025,wu2025hunyuanvideo15,openai2025sora2,seedance2026,hacohen2026ltx2,minimax2026h3} have made substantial progress in visual fidelity, motion coherence, and multimodal conditioning, with recent systems also supporting synchronized audio. Their ability to synthesize realistic scenes and dynamic content provides a foundation for generating interactive environments. Building on these advances, interactive world models~\cite{sun2025worldplay,gao2026infinite,abot2026,huang2026solarwm,alayaworld2026playable,echowm2026} generate subsequent observations in response to user or agent inputs, while broadening the range of interactions, extending generation horizons, and improving runtime efficiency. These advances bring us closer to explorable environments that evolve continuously and respond to their inhabitants in real time. They also motivate a further goal: recording and updating world state alongside immediate generative feedback, so that the consequences of actions persist and shape future events and observations. This would connect exploration, manipulation, and communication into a continuous experience grounded in history and shaped by action.

Existing approaches explore different aspects of interactive world modeling. Research on game-based simulation~\cite{guo2025mineworld,zhang2025matrixgame,li2025hunyuangamecraft,xiao2025worldmem} studies action-conditioned generation using gameplay data and virtual environments. Camera-controlled world models~\cite{he2025matrixgame2,sun2025worldplay,echowm2026,abot2026,huang2026solarwm,alayaworld2026playable} enable users to explore generated scenes through viewpoint changes and navigation. Text-controlled world models~\cite{gao2026infinite,deepmind2025genie3,runway2026gwmworlds2,mao2025yume15,tang2025hunyuangamecraft2,alayaworld2026playable} further allow users to influence character behavior and world events through natural-language instructions. These directions provide increasingly flexible interfaces for interaction within generated environments. An open challenge is to organize the resulting changes into an explicit shared world state that can be queried and updated, evolves according to defined rules, and preserves the consequences of past actions. This becomes particularly important when a world contains multiple actors, evolves over extended periods, or is revisited after an observer leaves. Such a system must maintain shared facts and history while generating each observer's local experience.

To address this challenge, we present a Generative World Simulation framework comprising \textbf{DAO} and \textbf{JING}. DAO is a computable world engine that maintains the shared world's state and computes its dynamics. JING is an Egocentric Interactive Experience Model that generates what each observer sees, hears, and experiences next. Actions change the world, and these changes shape the experiences and subsequent choices of humans and agents. This continuous loop connects perception, decision, action, and world evolution within a shared world that persists beyond any individual observer. Users can specify the scenes they explore, the objects they manipulate, and the characters they communicate with, continually influencing both the world and their own experience through interaction. The framework has three core capabilities:

\begin{itemize}[before={\interlinepenalty=10000}]
  \item \textbf{JING: Controllable first-person audiovisual experiences.} JING supports flexible combinations of character, object, and scene references, together with observation history and current actions, to generate continuous audiovisual content. Users explore environments through keyboard and mouse inputs and express manipulation and communication intents through text instructions, enabling first-person interaction across diverse settings and visual styles.

  \item \textbf{DAO: Computable and persistent shared world state.} DAO explicitly represents characters, objects, spaces, and their relationships, while maintaining world time and active events. The engine processes actions according to the current state and interaction rules, carries the resulting changes into subsequent states, and retains unchanged information. Past interactions can therefore influence future developments, and world state is preserved as observers change viewpoints, leave, or return.

  \item \textbf{Continuous closed-loop interaction in a shared world.} DAO and JING connect action consequences, local observations, and subsequent choices in an ongoing cycle. Humans and agents act on the audiovisual feedback they receive. The resulting changes enter the shared state and shape future experiences and interaction opportunities. From the same starting context, different choices can lead to different developments, allowing each experience to build on its history while responding to new actions.
\end{itemize}

We build JING by adapting MiniMax-H3 and training on diverse audiovisual data through a staged pipeline. Bidirectional teacher training introduces camera control, while causal autoregressive training with resampling forcing adapts the model to continuous interaction conditioned on generated histories. Memory adaptation and distillation support the use of historical information and efficient generation, respectively. DAO organizes shared world state through an entity graph, together with actor states, a world clock, and event mechanisms. It updates this state according to spatial constraints and interaction rules, with snapshots and event history supporting persistence. The world state is then converted into observer-specific conditions for JING's audiovisual generation. This design connects world-state computation with audiovisual experience generation. This report describes the data construction, model training, and world engine design, and examines generation quality, action responsiveness, state persistence, long-term consistency, and runtime efficiency.

Our long-term goal is a persistent shared world in which human and agent interactions have lasting consequences and world evolution continues beyond any individual observer's presence.

\section{Related Work}
\label{sec:related}

\subsection{Multimodal Video Generation}

Video generation has progressed from text- and image-conditioned visual synthesis toward synchronized audiovisual generation with diverse reference conditions. Wan~\cite{wan2025} and HunyuanVideo~1.5~\cite{wu2025hunyuanvideo15} advance visual quality, motion coherence, and condition adherence, providing a foundation for generating complex dynamic scenes. Joint audiovisual models, including SyncFlow~\cite{liu2024syncflow}, Sora~2~\cite{openai2025sora2}, and LTX-2~\cite{hacohen2026ltx2}, model visual and acoustic content together to produce temporally aligned outputs. Seedance~2.0~\cite{seedance2026} and MiniMax-H3~\cite{minimax2026h3} further extend multimodal reference conditioning and content control. Together, these advances enable models to represent scene appearance, character behavior, and accompanying sound. Building on this foundation, JING combines character, object, and scene references with observation history and ongoing action inputs to generate continuous, observer-centric interactive experiences.

\subsection{Interactive and Egocentric World Models}

Interactive world models predict future observations conditioned on actions, extending video generation to environments that users can continuously influence. MineWorld~\cite{guo2025mineworld}, Matrix-Game~\cite{zhang2025matrixgame}, and Hunyuan-GameCraft~\cite{li2025hunyuangamecraft} learn mappings from player actions to visual feedback in game environments. Matrix-Game~2.0~\cite{he2025matrixgame2}, WorldPlay~\cite{sun2025worldplay}, and AlayaWorld~\cite{alayaworld2026playable} further explore real-time control, streaming generation, and long-horizon interaction. LingBot-World~2.0~\cite{gao2026infinite} and Genie~3~\cite{deepmind2025genie3} broaden semantic control over character behavior and world events. In egocentric interaction, Hand2World~\cite{wang2026hand2world} studies gesture-driven object manipulation, EgoSim~\cite{hao2026egosim} uses an updatable 3D scene to support continued interaction, and EchoWM~\cite{echowm2026} combines navigation with video, environmental sound, and speech generation. JING focuses on integrating exploration, manipulation, and communication under flexible reference conditions, allowing these forms of interaction to unfold within a continuous first-person audiovisual experience.

\subsection{Long-Horizon Generation and Memory}

Long-horizon interaction requires models to use generated history within a bounded computational budget while limiting error accumulation during autoregressive rollouts. Diffusion Forcing~\cite{chen2024diffusionforcing} and Self Forcing~\cite{huang2025selfforcing} advance autoregressive video modeling through causal generation and training on model-generated histories, respectively. To manage growing context, FramePack~\cite{zhang2025framepack} compresses historical frame representations, while LongLive~\cite{yang2025longlive} combines streaming inference with early-frame retention. WorldMem~\cite{xiao2025worldmem} and WorldPlay~\cite{sun2025worldplay} improve long-term consistency through history retrieval and geometry-aware memory. OmniForcing~\cite{su2026omniforcing} extends streaming autoregressive generation to synchronized audiovisual content. For sampling efficiency, Distribution Matching Distillation (DMD)~\cite{yin2024dmd} and Velocity Moment Matching (VMM), introduced in EditStream~\cite{zhou2026editstream}, reduce the number of sampling steps through distillation. JING combines adaptation to generated histories, memory utilization, and few-step distillation to support sustained egocentric interaction.

\subsection{Persistent World State and Generative Simulation}

Explicit state modeling provides queryable and updatable representations of world facts, allowing world evolution to be governed by both state and rules. StatePlay~\cite{lin2026stateplay} jointly predicts visual content and game states to improve adherence to game mechanics. MASS~\cite{cai2026mass} separates shared-state updates from observation generation: a learned logic engine processes joint actions, and a rendering engine generates observations from the resulting shared state. For spatial and entity state modeling, EgoSim~\cite{hao2026egosim} maintains an updatable 3D scene, while Marionette~\cite{meng2026marionette} separates structured state prediction, geometric rendering, and appearance synthesis. Programmable World Model~\cite{huang2026programmable} further uses executable programs to maintain entity states and transition rules, compiling these states into spatial conditions for video generation. DAO focuses on ongoing character activities and event evolution under explicit interaction rules, translating local world state into observer-specific text, camera, and reference conditions for JING's first-person audiovisual generation.

\section{Data Pipeline}
\label{sec:data}
% \placeholder{Describe the actual data pipeline here. The following subsections reserve space for data sources, processing, and dataset construction.}

\begin{figure}[!htbp]
\centering
\includegraphics[width=\textwidth]{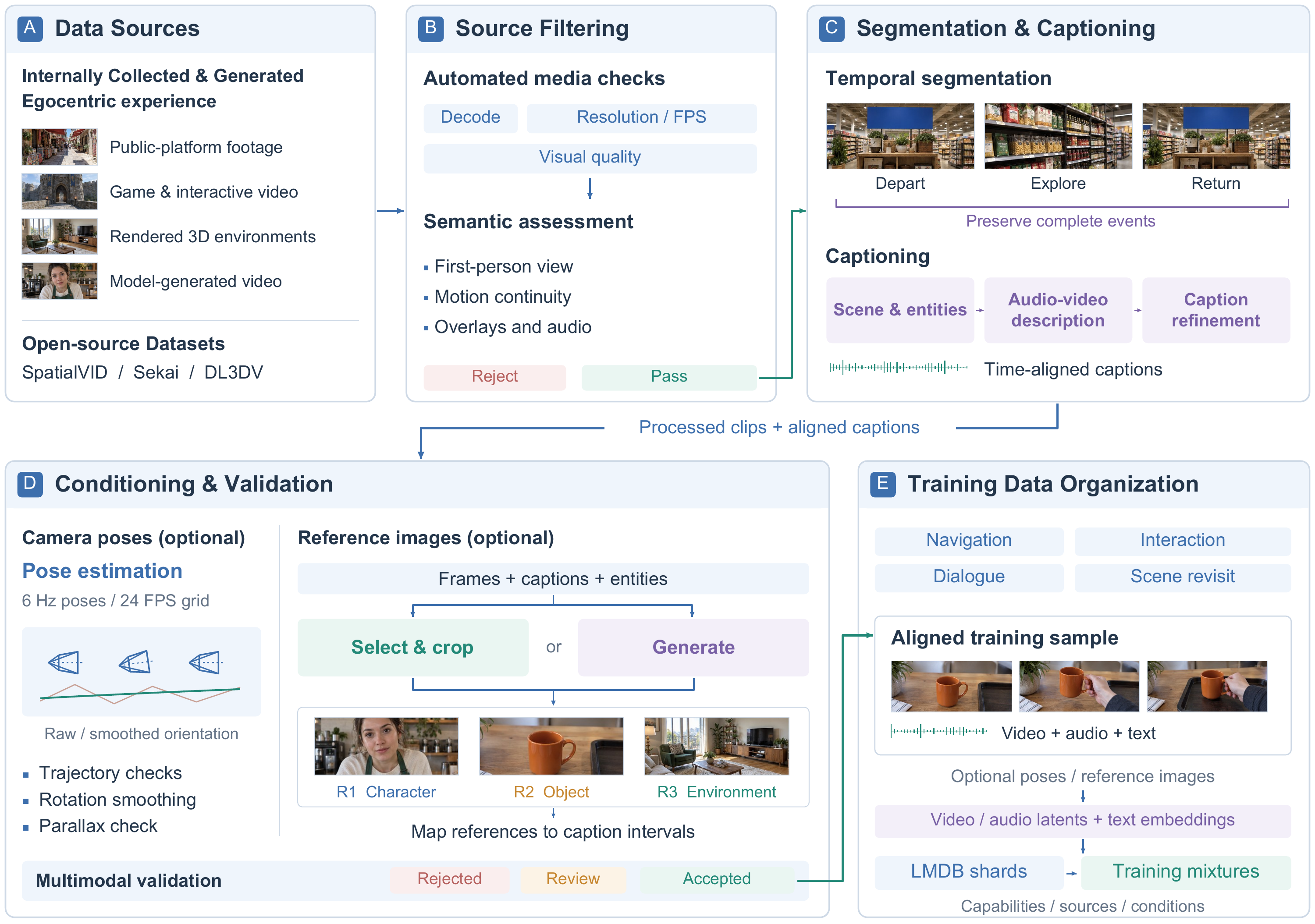}
\caption{\textbf{Data pipeline overview.} Internally collected and generated egocentric videos and open-source datasets undergo automated media checks, model-assisted semantic filtering, temporal segmentation, and captioning. Optional camera trajectories and extracted or generated reference images are aligned with the processed clips and validated before accepted samples are organized by capability and conditioning configuration and packaged into latent representations and LMDB shards for training.}
\label{fig:data-pipeline-overview}
\end{figure}

Our data pipeline converts diverse video sources into aligned multimodal training samples for JING. As shown in Figure~\ref{fig:data-pipeline-overview}, the pipeline spans data acquisition, source-level filtering, temporal segmentation, captioning, conditioning, and training data organization. Its design emphasizes temporal continuity across navigation, scene revisits, interactions, and dialogue, together with consistent alignment among video, audio, text, and optional camera or reference-image conditions. The following subsections describe the data sources, processing and quality-control procedures, and organization of the resulting training corpus.

\subsection{Data Sources}
% Document the origin and scope of each data source, including the supported domains and modalities. State the collection procedure, dataset version, and inclusion criteria. Report dataset sizes only after the collection and filtering stages have been verified.

We construct our training corpus from two complementary sources: internally collected and generated data and open-source datasets. Our data effort centers on egocentric experience: how an environment unfolds from an observer’s first-person perspective as they move through scenes, revisit locations, interact with surrounding entities, and engage in conversations. These experiences guide the selection and construction of our training material.

\paragraph{Internally Collected and Generated Data.}

Our internal corpus focuses on egocentric navigation, scene revisits, interaction, and dialogue. We collect first-person videos from a range of publicly accessible platforms, combining broad scene-exploration footage with targeted collections of specific environments and scenarios. These videos capture how scenes unfold as an observer moves through them, encounters nearby entities, and returns to previously observed locations.

We complement this material with game and interactive-video footage acquired through dedicated collection efforts by our team and external partners. These sources expand our coverage of navigation, interactions, and conversations within evolving scenes. We further supplement the corpus with videos rendered in controllable 3D environments and model-generated videos depicting interactions and dialogue, allowing us to construct additional experiences around selected scenes and events. Across these acquisition and generation routes, we prioritize the observer’s first-person perspective and the temporal context of each experience, including continuity across camera movements, scene revisits, and interactions.

\paragraph{Open-source Datasets.}

We additionally incorporate publicly released datasets, including SpatialVID~\cite{wang2026spatialvid}, Sekai~\cite{li2025sekai}, DL3DV~\cite{ling2024dl3dv}. Selected subsets complement our internal collection with additional scene content and camera-motion patterns, broadening the visual and spatial diversity of the training corpus. Together with our targeted collection efforts, these resources provide video material depicting how environments unfold across changing viewpoints and observer movements. 
% We organize public data under their original dataset identities and retain source information for derived training samples. This preserves provenance through subsequent processing and allows individual datasets and their processed versions to be inspected separately. 
% The processing and quality-control procedures used to prepare these videos for training are described in Section 3.2.

\subsection{Processing and Quality Control}
% Explain how raw inputs are converted into training examples. Describe preprocessing, quality assessment, duplicate removal, and annotation procedures. Define the acceptance criteria for each stage and distinguish automated checks from human review.

% Record how examples are tracked through the pipeline so that the final dataset can be traced back to its source. Discuss any known coverage gaps or limitations introduced by the selection process.

Our processing pipeline converts collected videos into training samples with aligned video, audio, captions, and optional conditioning signals. We apply common quality criteria across sources while adapting segmentation and annotation to the temporal structure of navigation, scene revisits, interactions, and dialogue. The objective is to preserve continuous egocentric experiences and provide consistent supervision for training.

\paragraph{Source-level filtering.}
We first apply automated checks to media metadata and decoding results, and compute visual-quality metrics on sampled, low-resolution video segments. These checks assess resolution, frame rate, decoding reliability, sharpness, exposure, compression artifacts, and frozen frames. Predefined rules exclude videos with severe defects, while cases requiring contextual interpretation, such as moderate motion blur or potential shot transitions, are retained for further assessment.

Videos that pass this stage undergo semantic assessment using VLM. The model examines frames sampled across the video timeline, short windows sampled at higher frame rates, and available audio to assess first-person perspective, motion continuity, shot transitions, visual overlays, and unrelated background music. The viewpoint criterion requires the camera to represent the observer’s direct experience, without requiring visible hands or other body parts. Motion is evaluated in context: natural walking movements and coherent head turns are acceptable, while unexplained jumps or interruptions that break temporal continuity are grounds for rejection. The model’s structured outputs are checked against predefined acceptance rules to determine whether each source video can proceed to subsequent processing.

\paragraph{Temporal segmentation.}
After filtering, we use Qwen3.6~\cite{qwen2026qwen36} to identify candidate segment boundaries from sampled video frames. These boundaries are adjusted according to the activity being captured. For scene revisits, segmentation aims to keep the departure and return within the same sequence. For interactions and dialogue, it aims to preserve complete operations and conversational exchanges. Each segment retains its source identifier and original time interval, allowing the processed video to be traced back to its source.

\begin{figure}[!t]
\centering
\includegraphics[width=0.9\textwidth,keepaspectratio]{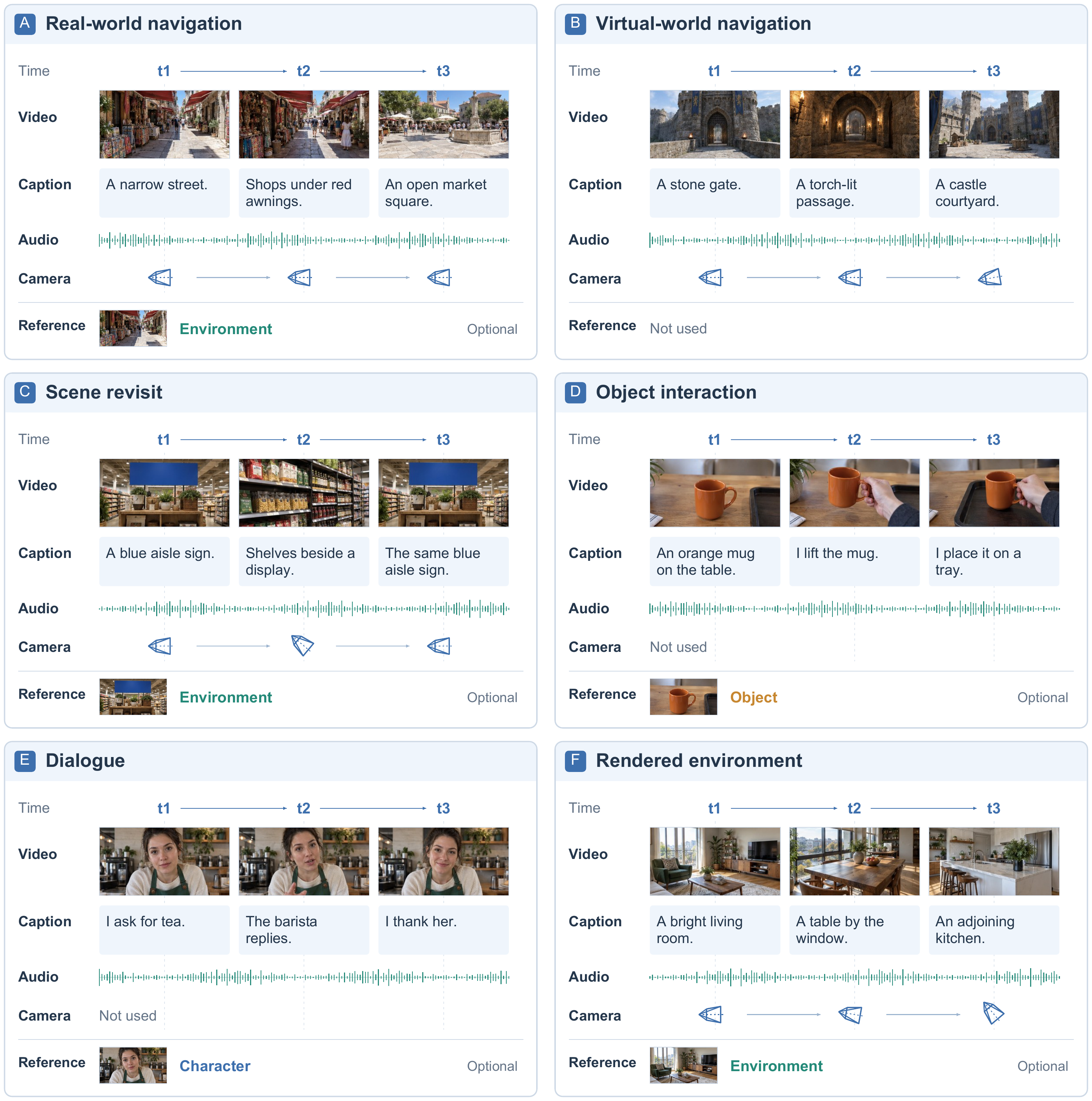}
\caption{\textbf{Diverse aligned training samples.} Illustrative samples covering real-world and virtual-world navigation, scene revisits, object interaction, dialogue, and rendered environments. Within each panel, video frames, captions, audio, and available camera poses follow a common temporal axis, with optional reference images for character, object, or scene conditioning.}
\label{fig:aligned-training-samples}
\end{figure}

Processed clips are packaged into sharded LMDB datasets containing video and audio latents, caption-aligned text embeddings, and applicable camera or reference-image conditions. This unified format supports data from different processing pipelines and enables variants with and without reference-image conditioning to share the same audiovisual content. Each record retains its source, temporal interval, and conditioning settings, preserving alignment and traceability across supervision variants.
Figure~\ref{fig:aligned-training-samples} illustrates the temporal alignment and optional conditioning of multimodal samples across diverse egocentric experiences.

\paragraph{Video captioning.}
We first identify recurring entities and scene context to maintain consistency across segments, then generate captions describing visible events, actions, speech, and relevant sounds. These descriptions are refined into training captions with a first-person perspective. Captions paired with camera trajectories avoid redundant camera-motion descriptions, while those without explicit camera conditioning include the relevant camera movements.

\paragraph{Conditioning and validation.}
We construct camera and reference-image conditions for selected samples. For camera conditioning, ViPE~\cite{huang2025vipe} estimates camera intrinsics and six-degree-of-freedom poses from processed clips. The resulting trajectories are smoothed and filtered to reduce jitter and exclude unreliable estimates, while maintaining temporal alignment with the video.

Reference images specify characters, objects, or scenes and are obtained by selecting and cropping targets from video frames or generating images from relevant frames and prompts. Each reference is linked to the corresponding caption intervals to align visual conditions with the described content.

Before training, we validate audiovisual synchronization and consistency among media, captions, metadata, and conditioning signals. Automated checks are complemented by inspection of sampled examples, and only samples that pass validation are retained.

\subsection{Training Data Organization}

After processing and quality control, we organize the accepted data according to the egocentric capabilities they support, including navigation, interaction, and dialogue. Each sample retains its primary capability and relevant secondary attributes, allowing a sequence to capture multiple aspects of an experience. The availability of camera trajectories and reference images further determines the conditioning configurations supported by each sample.

\section{Methods}
\label{sec:methods}

\subsection{JING-Experience Model}
\label{sec:experience-model}

\begin{figure}[!htb]
  \centering
  \includegraphics[width=\linewidth]{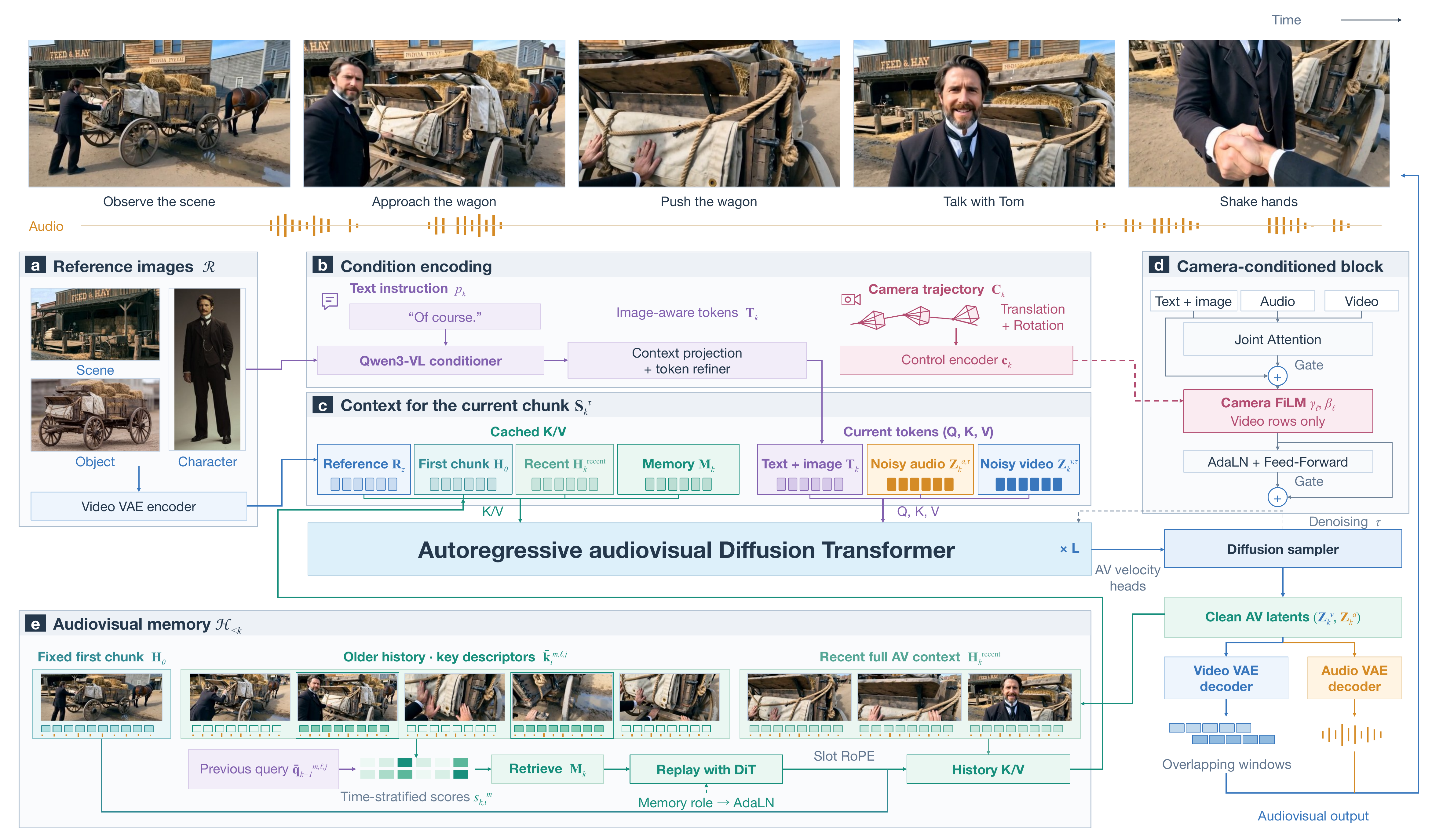}
  \caption{\textbf{JING architecture.} (a) Reference images provide visual conditions through the video VAE encoder and (b) the Qwen3-VL conditioner, which also encodes text instructions. Camera trajectories are processed by a separate control encoder. (c) Current tokens attend to cached keys and values (K/V) from reference and historical context. (d) Camera features modulate video tokens within each Transformer block. (e) Retrieved audiovisual memories supplement the persistent first chunk and recent context. The diffusion sampler produces clean audiovisual latents, which the video and audio VAE decoders convert into synchronized output.}
  \label{fig:jing-architecture}
\end{figure}

JING is an Egocentric Interactive Experience Model built on MiniMax-H3~\cite{minimax2026h3} that generates continuous audiovisual content autoregressively. As shown in Figure~\ref{fig:jing-architecture}, at generation step $k$, the model takes a set of reference images $\mathcal R$, a text instruction $p_k$, camera trajectory $\mathbf C_k$, and the previously generated audiovisual history $\mathcal H_{<k}$, and predicts the next video and audio latent chunk:
\begin{equation}
  (\mathbf Z_k^v,\mathbf Z_k^a)
  \sim
  P_\theta\!\left(
    \,\cdot\mid\mathcal R,p_k,\mathbf C_k,\mathcal H_{<k}
  \right),
  \label{eq:jing-generation}
\end{equation}
where $v$ and $a$ denote video and audio, respectively. The corresponding VAEs decode these latents into synchronized frames and audio, and the generated chunk becomes part of the history for subsequent generation.
Reference images are jointly encoded with text by Qwen3-VL~\cite{bai2025qwen3vl} into semantic conditioning tokens $\mathbf{T}_k$ and separately encoded by the video VAE into reference tokens $\mathbf{R}_z$, providing semantic conditioning and visual reference information, respectively.

% Text instructions specify content, actions, and conversational intent, while camera trajectory describes changes in the observer's viewpoint. Together, these inputs allow the model to extend the first-person experience as interaction continues.

To further support streaming generation, we preserve the temporal order of the latent sequence and reorganize H3’s temporal partition from $(5,5,\ldots,2)$ to $(2,5,5,\ldots)$. The initial $2+5$ video latent frames form an initialization chunk, followed by chunks of five latent frames each. This layout aligns generation boundaries with the VAE’s overlapping decoding windows, enabling incremental decoding as new chunks become available.

Camera conditions enter the Transformer through a dedicated control encoder. Relative translations and rotations between consecutive camera poses form a six-dimensional motion representation, which the encoder converts into control features $\mathbf c_k$ aligned with the video tokens in space and time. In Transformer block $\ell$, these features modulate the video hidden states through feature-wise linear modulation (FiLM):
\begin{equation}
  \widetilde{\mathbf h}_{k,v}^{\,\ell}
  =
  \left(1+\gamma_\ell(\mathbf c_k)\right)
  \odot\mathbf h_{k,v}^{\,\ell}
  +\beta_\ell(\mathbf c_k),
  \label{eq:jing-camera-film}
\end{equation}
where $\gamma_\ell$ and $\beta_\ell$ predict scale and shift, respectively, and $\odot$ denotes element-wise multiplication. The modulation acts on camera-conditioned video tokens after the attention residual update and before the subsequent normalization and feed-forward network. Text and audio tokens do not receive this direct modulation, but exchange information with visual features through shared attention. This design introduces camera control while retaining a unified audiovisual backbone.

The attention context for the current chunk combines reference conditions, history at different temporal scales, and the content being generated. After projection into a shared hidden space, its logical composition is
\begin{equation}
  \mathbf S_k^\tau=
  \left[
    \underbrace{
      \mathbf R_z;\,\mathbf H_0;\,
      \mathbf H_k^{\mathrm{recent}};\,\mathbf M_k
    }_{\text{reference and historical context}}
    ;\,
    \underbrace{
      \mathbf T_k;\,\mathbf Z_k^{a,\tau};\,\mathbf Z_k^{v,\tau}
    }_{\text{current chunk}}
  \right],
  \label{eq:jing-context}
\end{equation}
where $\tau$ denotes the noise level during denoising, $\mathbf H_0$ is the fixed first generated chunk, $\mathbf H_k^{\mathrm{recent}}$ contains the full audiovisual context from recent chunks, and $\mathbf M_k$ represents memories retrieved from older history. History terms are empty until the corresponding chunks have been generated. Semantic, audio, and video tokens interact through joint attention in the shared Transformer. Attention is joint within each chunk, while access across chunks is causal. 
%At inference time, reference and historical context provide cached keys and values (K/V), and current tokens produce queries, keys, and values (Q/K/V). This organization avoids recomputing the full history at every denoising step. 
The fixed first chunk provides an initial anchor, recent context supports local continuity, and retrieved memories supply information beyond the current window.

More specifically, JING maintains an audiovisual memory archive to access history beyond the recent window. Retrieval units are video latent frames and contiguous groups of stereo audio latents. After each chunk is generated, the model extracts compact query and key descriptors from the Q/K representations of selected attention layers. At inference time, the query descriptor from the last completed chunk combines information from text and the corresponding modality to score earlier history units:
\begin{equation}
  s_{k,i}^{m}
  =
  \operatorname{Agg}_{\ell,j}
  \left(
    \frac{
      \left\langle
        \bar{\mathbf q}_{k-1}^{m,\ell,j},
        \bar{\mathbf k}_{i}^{m,\ell,j}
      \right\rangle
    }{\sqrt{d_{\mathrm{head}}}}
  \right),
  \qquad m\in\{v,a\}.
  \label{eq:jing-memory-retrieval}
\end{equation}
Here, $\bar{\mathbf q}$ and $\bar{\mathbf k}$ denote pooled query and key descriptors, respectively. The indices $i$, $\ell$, and $j$ identify historical candidates, attention layers, and attention heads, while $d_{\mathrm{head}}$ denotes the feature dimension of each head. The operator $\operatorname{Agg}$ standardizes candidate scores before aggregating them across heads and layers. The candidate pool excludes the retained first chunk and recent context. Candidates are divided into temporal intervals, and relevant units are selected from each interval to balance relevance with historical coverage. The selected units are replayed through the Transformer with dedicated RoPE position slots and AdaLN memory-role embeddings to obtain memory K/V for the current context. Each newly generated chunk is then added to the archive for future retrieval. By separately limiting the recent window and audiovisual memory budgets, the model bounds the historical context used by current attention while retaining access to earlier experiences.

\subsection{JING-Model Training}
\label{sec:training}

Using the audiovisual data described in Section~\ref{sec:data}, we train MiniMax-H3 in five stages to introduce camera control, enable causal generation, incorporate long-term memory, and reduce sampling steps. Together, these stages adapt the pretrained model for continuous, real-time first-person interaction.

We first train a bidirectional teacher with the camera-control module described in Section~\ref{sec:experience-model}. Camera-motion features modulate the Transformer alongside the existing text and visual conditions. We retain bidirectional attention and jointly optimize model parameters and the control module, allowing the model to learn camera-conditioned motion from complete audiovisual clips. This lightweight extension introduces camera control while retaining the pretrained backbone architecture. The resulting model serves as the teacher for subsequent distillation.

We then train a causal autoregressive model to generate audiovisual chunks conditioned on preceding content and current user inputs. We augment teacher-forcing training~\cite{williams1989teacherforcing} with resampling forcing~\cite{guo2025resamplingforcing}: historical segments are perturbed with noise, reconstructed by the model, and reused as context for predicting subsequent content. These reconstructed histories reduce reliance on error-free context and help bridge the gap between clean training histories and model-generated histories at inference. Camera and text conditions can be updated at chunk boundaries, allowing generation to respond to ongoing interaction.

Next, we introduce the audiovisual memory described in Section~\ref{sec:experience-model} and train the model to combine the fixed initial context, a recent-history window, and retrieved memories. During adaptation, we vary the recent-context length, memory selection, and available memory capacity so that the model learns to use different combinations of historical information. This provides access to earlier scenes and events within a bounded context budget: recent content supports local continuity, while retrieved memories supply cues from beyond the recent window.

To reduce denoising steps, we initialize a few-step student with Velocity Moment Matching (VMM)~\cite{zhou2026editstream}. VMM matches conditional velocity moments at intermediate noisy states reached during the student's denoising process. A frozen teacher and a trainable auxiliary network provide velocity estimates that guide the student's updates. This objective aligns with the backbone's flow-matching parameterization and supervises states encountered during few-step sampling, providing an initialization for subsequent distribution matching.

Finally, we refine the student with Distribution Matching Distillation (DMD)~\cite{yin2024dmd}. The student generates audiovisual sequences conditioned on its own generated history. A frozen teacher and a score model trained on student-generated samples guide the student's output distribution toward the teacher's. Building on VMM initialization, this stage aims to recover quality lost when reducing sampling steps and improve autoregressive generation. The full training pipeline combines camera control, adaptation to imperfect histories, and memory utilization with efficient few-step sampling.

\subsection{DAO-World Engine}
\label{sec:dao-world-engine}

\begin{figure}[!tb]
  \centering
  \includegraphics[width=\linewidth]{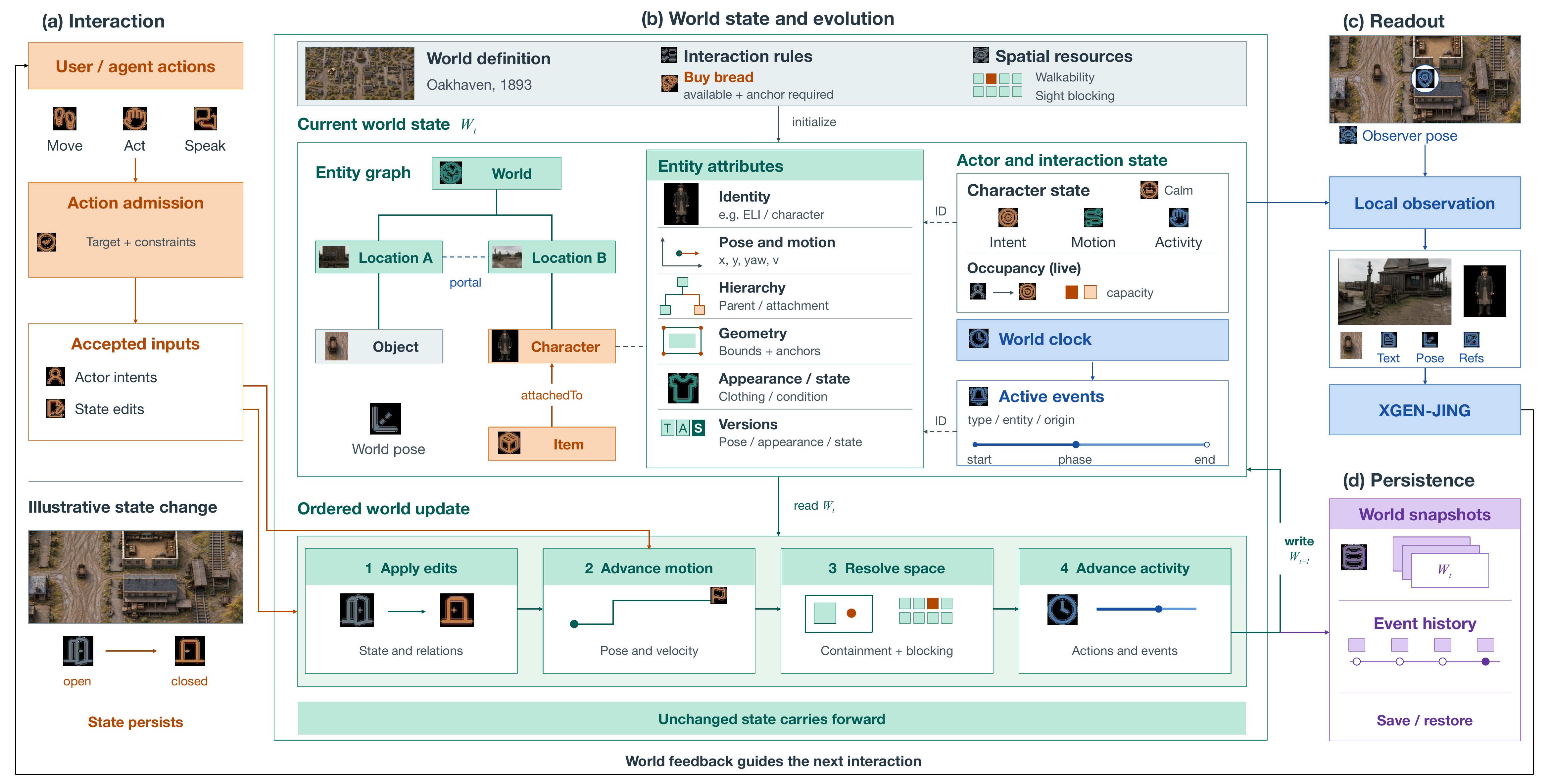}
  \caption{\textbf{DAO world engine.} (a) User and agent actions pass through action admission to produce actor intents and state edits. (b) A world definition initializes a shared state comprising an entity graph, entity attributes, actor states, a clock, and active events. An ordered update cycle transforms $W_t$ into $W_{t+1}$ by applying edits, advancing motion, resolving spatial constraints, and advancing activities, while carrying unchanged state forward. (c) Local readout combines the world state with an observer's pose to provide observations for an experience interface. (d) World snapshots and event history preserve state across interactions. }
  \label{fig:dao-architecture}
\end{figure}

DAO is a computable world engine that maintains explicit shared state and preserves the consequences of user and agent actions. As shown in Figure~\ref{fig:dao-architecture}, a world definition specifies the spatial layout, initial entities, interaction rules, and spatial constraints. State updates are separate from observation readout, allowing different observers to derive their experiences from the same underlying world.

% Keep the architecture figure on a text page.

\begin{samepage}
We represent the shared world state at simulation step $t$ as
\begin{equation}
  W_t = (G_t, B_t, E_t, \omega_t),
  \label{eq:dao-state}
\end{equation}
\end{samepage}

where the entity graph $G_t$ represents places, characters, and objects through their poses, appearance, semantic attributes, and spatial and attachment relations. Actor and interaction state $B_t$ records intentions, motion, activities, and attributes such as emotions, needs, and object use. The active event set $E_t$ represents ongoing events, while the world clock $\omega_t$ provides a shared time reference. Together, these components provide a queryable representation of the world for action execution and observation generation.

Inputs first pass through action admission, which resolves targets and checks interaction rules and state constraints, including reachability and target availability. Accepted inputs produce actor intents or queued state edits; inputs that violate these constraints can be rejected. Given admitted inputs $\widehat{\mathcal U}_t$ and world definition $\mathcal D$, state evolution is expressed as

\begin{equation}
  W_{t+1} = F_{\mathcal D}(W_t, \widehat{\mathcal U}_t),
  \label{eq:dao-transition}
\end{equation}

where $F_{\mathcal D}$ is the update process determined by the world configuration and rules. Each step applies queued edits, advances actor motion, updates spatial membership and dynamic obstacles, and then advances activities and events. Unchanged state carries forward, so earlier interactions remain part of the world and can affect subsequent actions and observations.

\begin{table}[H]
\centering
\caption[Quantitative comparison on WBench]{\textbf{Quantitative comparison on WBench~\citep{ying2026wbench}.}\protect\footnotemark Ranks refer to the comparison pools of 15 Full and 39 Navigation entries. Bold and underlined values indicate the best and second-best results among the models shown within each split; shaded rows denote our models.}
\label{tab:wbench}
\begingroup
\setlength{\tabcolsep}{3pt}
\renewcommand{\arraystretch}{0.9}
\fontsize{9}{11}\selectfont
\begin{tabularx}{\textwidth}{@{}rX*{6}{r}@{}}
\toprule
Rank & Model & Avg. & \shortstack[r]{Video\\quality} & \shortstack[r]{Setting\\adherence} & \shortstack[r]{Interaction\\adherence} & Consistency & Physics\\
\midrule

\multicolumn{8}{@{}l}{\textbf{Full (289 cases)}}\\
\addlinespace[3pt]
\rowcolor{black!5}
1 & \textbf{JING (Bidirectional)} (ours) & \textbf{81.0} & 80.9 & 88.5 & \underline{76.7} & \underline{85.7} & \textbf{73.3}\\
\rowcolor{black!5}
2 & \textbf{JING (4-step AR)} (ours) & \underline{79.9} & 79.8 & 88.9 & \textbf{79.5} & 81.4 & 69.9\\
3 & Kling 3.0~\citep{kling2026video3} & 79.4 & 80.0 & \underline{91.0} & 73.1 & 83.9 & 69.2\\
4 & Wan 2.7~\citep{alibaba2026wan27} & 78.4 & \underline{81.0} & \textbf{91.4} & 72.1 & 75.8 & \underline{71.6}\\
5 & Seedance 1.5~\citep{seedance2025pro15} & 76.2 & \textbf{81.9} & 82.9 & 68.3 & 79.9 & 68.2\\
6 & HY-Video 1.5~\citep{wu2025hunyuanvideo15} & 74.3 & 76.6 & 85.6 & 54.7 & \textbf{87.5} & 67.1\\
\midrule
\multicolumn{8}{@{}l}{\textbf{Navigation (158 cases)}}\\
\addlinespace[3pt]
1 & Alaya-EVOKE-Turbo~\citep{yin2026alayaevoke} & \textbf{82.0} & \underline{81.9} & 82.1 & 83.9 & 88.1 & \textbf{74.0}\\
\rowcolor{black!5}
2 & \textbf{JING (Bidirectional)} (ours) & \underline{81.9} & \textbf{82.4} & 90.3 & 75.6 & 88.1 & 73.3\\
3 & JoyAI-Echo-1.5 (Bidirectional)~\citep{echowm2026} & 81.6 & 81.5 & 79.4 & \underline{86.6} & \textbf{89.8} & 70.6\\
4 & JoyAI-Echo-1.5 (4-step AR)~\citep{echowm2026} & 81.0 & 81.1 & 77.5 & \textbf{87.9} & 88.3 & 70.1\\
5 & Zing-0.5~\citep{chen2026zing05} & 81.0 & 80.6 & 77.8 & 84.2 & \underline{88.5} & \underline{73.8}\\
\rowcolor{black!5}
6 & \textbf{JING (4-step AR)} (ours) & 81.0 & 81.3 & \textbf{91.6} & 78.7 & 83.7 & 69.8\\
7 & HiDream-O1-World~\citep{hidream2026o1world} & 80.9 & 81.0 & 82.2 & 80.0 & 88.0 & 73.3\\
10 & Kling 3.0~\citep{kling2026video3} & 79.0 & 81.4 & 91.0 & 69.4 & 83.7 & 69.3\\
13 & Wan 2.7~\citep{alibaba2026wan27} & 78.1 & 81.5 & \underline{91.4} & 64.4 & 81.6 & 71.8\\
29 & Genie 3~\citep{deepmind2025genie3} & 73.9 & 75.2 & 72.5 & 73.4 & 82.6 & 65.7\\
\bottomrule
\end{tabularx}
\endgroup
\end{table}
\footnotetext{Data as of September 17, 2026. \url{https://meituan-longcat.github.io/WBench/}}

DAO derives local observations from the shared state and observer pose, selecting relevant entities according to sensing range, field of view, and occlusion. These observations provide local entity states, events, and environmental information. Together with camera trajectories and character, object, and scene references, they condition JING's generation of the next first-person audiovisual experience. Different observers therefore receive local views grounded in the same shared state, and their subsequent actions re-enter the update cycle.

World snapshots and event history support persistence across sessions and process restarts. Snapshots store the state needed to resume simulation, while event history records past interactions. On restoration, DAO reconstructs the world from the saved snapshot and world definition, allowing subsequent actions and observations to reflect the preserved state.

\section{Model Performance}
\label{sec:performance}

We evaluate JING with other video generation and interactive models on WBench~\cite{ying2026wbench} and demonstrate its capabilities in reference conditioning, spatial exploration, object manipulation, character communication, branching experiences, and embodied interaction. We further examine DAO’s ability to maintain evolving world states and preserve the consequences of past actions. Together, these results demonstrate XGEN’s capacity to support continuous interactive experiences grounded in a persistent world.

\subsection{Quantitative Results}
\label{sec:quantitative}

As shown in Table~\ref{tab:wbench}, we evaluate JING on the Navigation and Full splits of WBench~\citep{ying2026wbench}. Navigation assesses interactive world generation across video quality, setting adherence, interaction adherence, consistency, and physical plausibility. Full additionally covers text-driven subject actions, event editing, and perspective switching, extending evaluation beyond navigation. On the Full evaluation set, our bidirectional and four-step autoregressive (AR) variants achieve average scores of 81.0 and 79.9, ranking first and second among 15 comparison entries. On the Navigation subset, they score 81.9 and 81.0, ranking second and sixth among 39 entries. These results show that our training and distillation pipeline transfers the bidirectional model to a four-step causal generator while largely preserving its overall performance.

% Let figures share pages with the surrounding discussion.
\begingroup
\renewcommand{\topfraction}{0.95}
\renewcommand{\bottomfraction}{0.95}
\renewcommand{\textfraction}{0.05}
\setlength{\floatsep}{7pt}
\setlength{\textfloatsep}{7pt}
\setlength{\intextsep}{2pt}
\captionsetup[figure]{skip=3pt}
\titlespacing*{\paragraph}{0pt}{5pt}{.65em}
\subsection{Qualitative Results}
\label{sec:qualitative}

\paragraph{Flexible reference conditioning.}
Figure~\ref{fig:perf-reference} illustrates how JING uses different reference combinations to specify the content of an experience. The generated sequences retain recognizable scene features, such as the clock tower and bakery interior, while incorporating the supplied character and object appearances where provided. These examples show how references constrain complementary aspects of generation: the environment to explore, the person to encounter, and the object to interact with.

\begin{figure}[H]
\centering
\includegraphics[width=.80\linewidth]{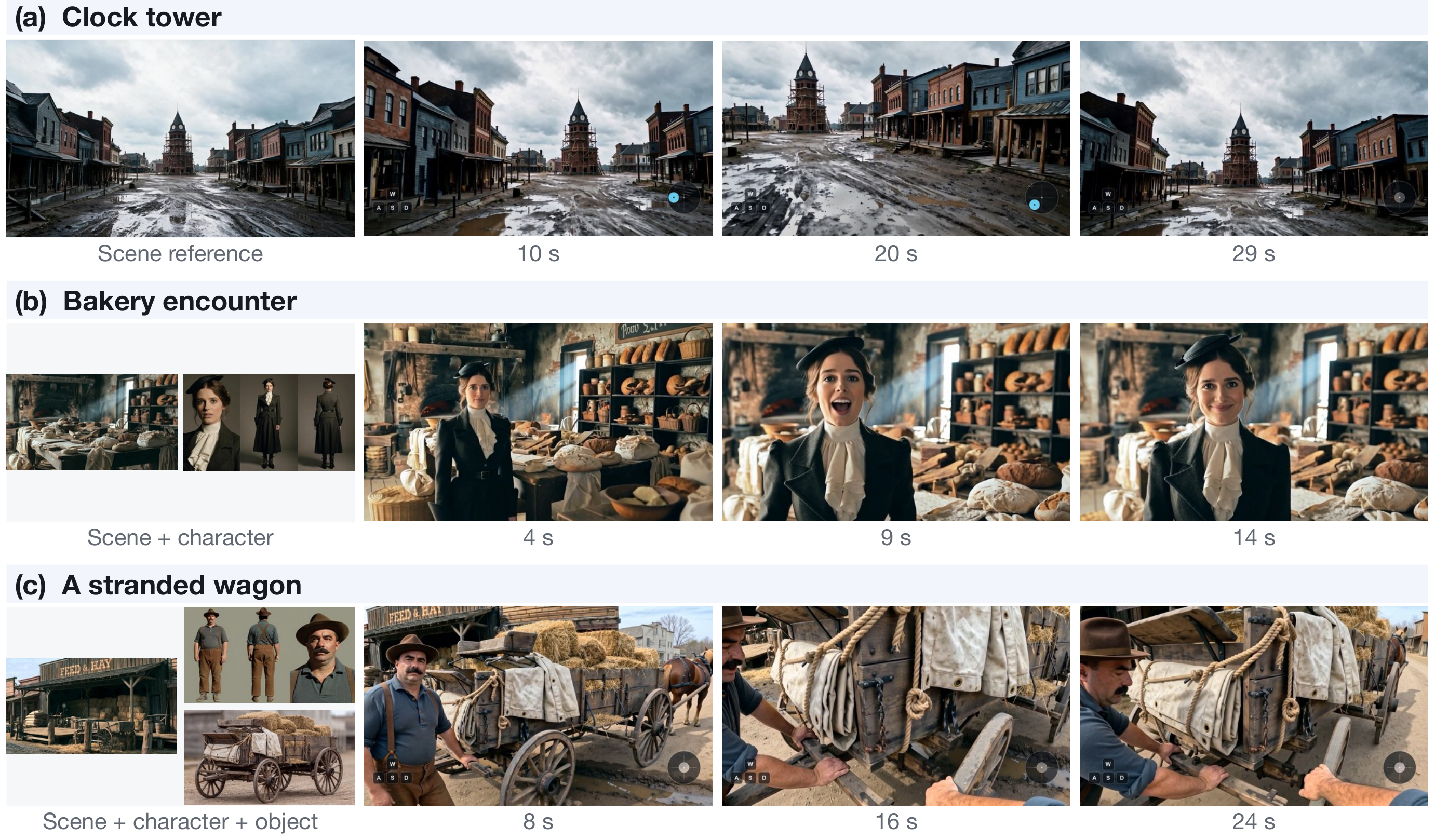}
\caption{\textbf{Reference-conditioned generation.} Input references and sampled outputs for three configurations: scene only (a), scene and character (b), and scene, character, and object (c).}
\label{fig:perf-reference}
\end{figure}

\paragraph{Egocentric exploration.}
JING translates navigation inputs into changing first-person views across different environments and observer scales (Figure~\ref{fig:perf-navigation}). Moving through an archway reveals a street, turning near a shrine exposes surrounding paths, and navigating at beetle scale reveals the space beneath leaves. Nearby surfaces and landmarks remain recognizable across successive views, illustrating local visual continuity as exploration reveals previously unseen content.

\begin{figure}[H]
\centering
\includegraphics[width=.80\linewidth]{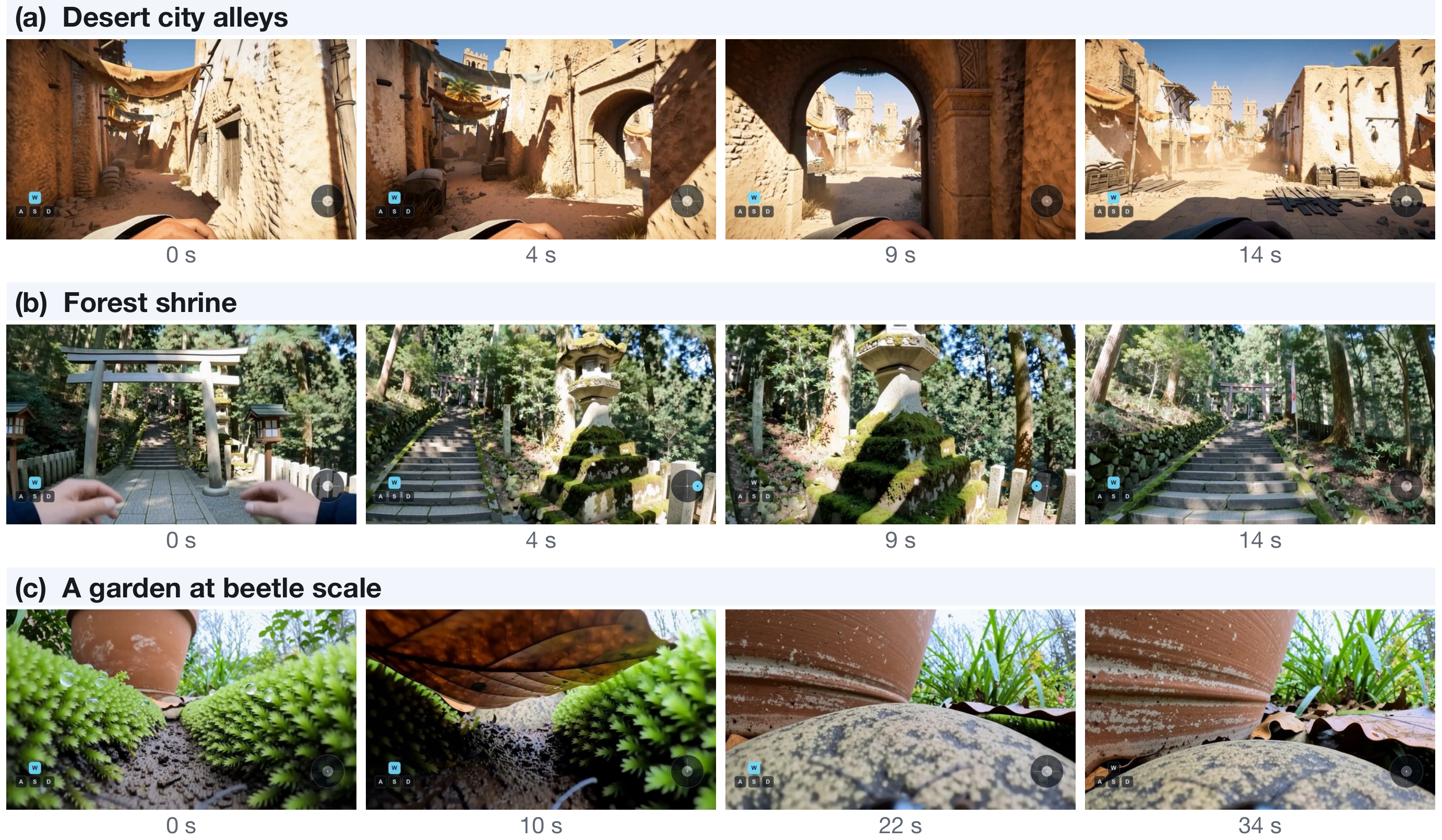}
\caption{\textbf{Egocentric exploration.} Sampled frames from navigation through desert-city alleys (a), a forest shrine (b), and a garden at beetle scale (c). Overlaid controls indicate user inputs during generation.}
\label{fig:perf-navigation}
\end{figure}

\paragraph{Object manipulation.}
Beyond viewpoint control, JING generates interactions that change nearby objects (Figure~\ref{fig:perf-action}). The sequences depict tool contact, ignition, and release, linking first-person actions to visible outcomes such as a burning woodpile or an ascending lantern. They illustrate action-conditioned changes in object state and motion across realistic and stylized settings.

\begin{figure}[H]
\centering
\includegraphics[width=.80\linewidth]{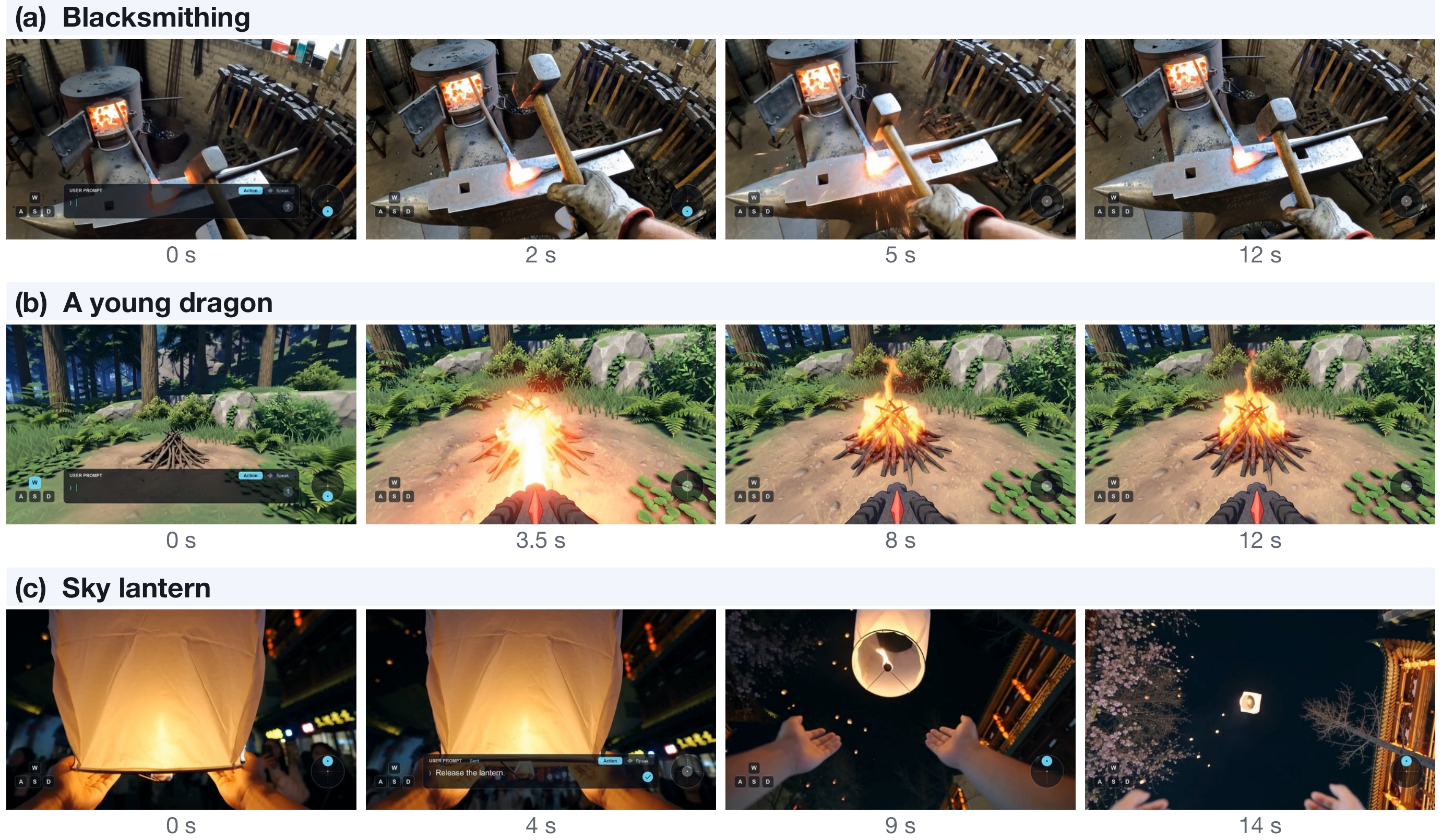}
\caption{\textbf{Object manipulation.} Sampled frames showing a hammer striking heated metal (a), a young dragon igniting a woodpile (b), and hands releasing a sky lantern (c).}
\label{fig:perf-action}
\end{figure}

\paragraph{Character communication.}
JING extends first-person interaction to character communication through generated speech and visual behavior (Figure~\ref{fig:perf-communication}). Across the illustrated languages and visual styles, characters address the observer with changes in gaze, expression, and posture while remaining within the ongoing scene.

\begin{figure}[H]
\centering
\includegraphics[width=.80\linewidth]{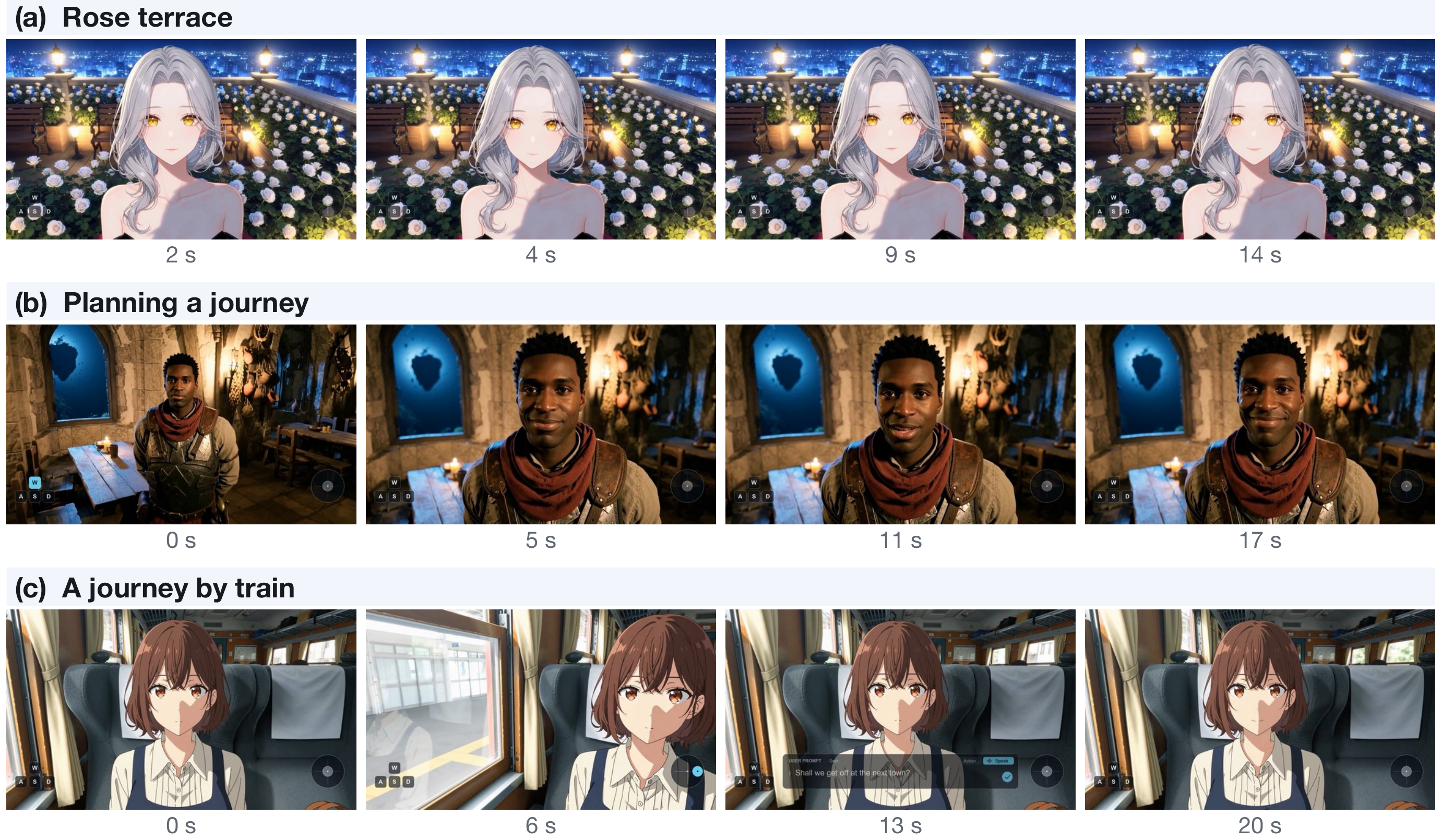}
\caption{\textbf{Character communication.} Sampled frames from conversations in Chinese on a rose terrace (a), English in a fantasy setting (b), and Japanese on a train (c). The corresponding videos include generated speech.}
\label{fig:perf-communication}
\end{figure}

\paragraph{Action-conditioned branching.}
Different user inputs can guide distinct continuations from a shared visual context (Figure~\ref{fig:perf-choices}). In the fountain scene, viewpoint control reveals coins in the water, a manipulation instruction produces a coin-toss gesture, and a request to meet a fairy introduces a character encounter. The common opening connects these branches, illustrating how exploration, manipulation, and communication offer different ways to develop the same initial situation.

\begin{figure}[H]
\centering
\includegraphics[width=.80\linewidth]{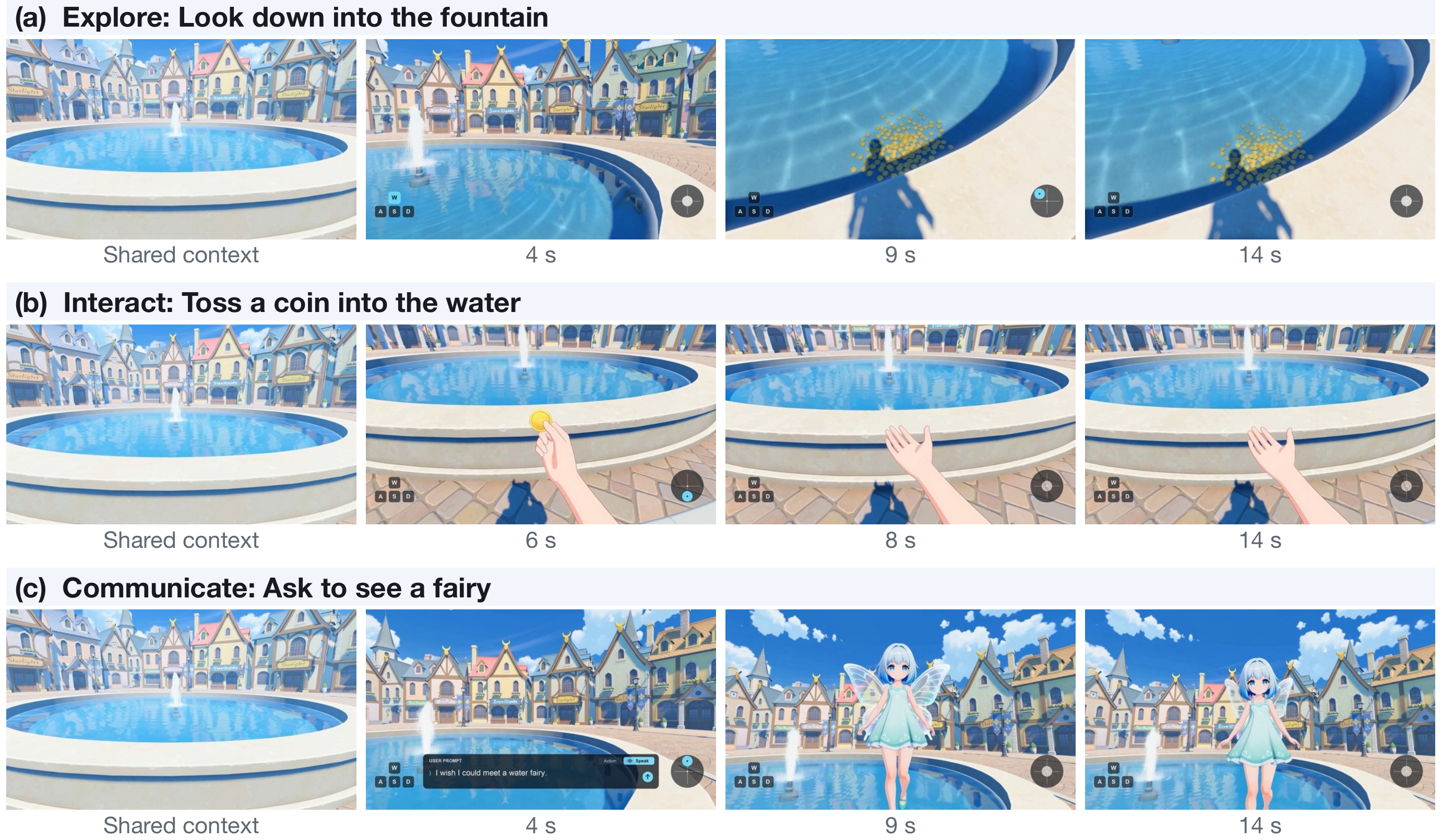}
\caption{\textbf{Action-conditioned branching.} A shared fountain opening leads to three continuations: looking into the water (a), tossing a coin (b), and meeting a fairy (c).}
\label{fig:perf-choices}
\end{figure}

\paragraph{Reference-conditioned branching.}
Reference selection also shapes how an experience unfolds (Figure~\ref{fig:perf-reference-choices}). Starting from a common bakery opening, the three continuations follow a similar entrance and greeting sequence, but the encountered character changes with the supplied reference. The examples illustrate control over who appears in an interaction while retaining the initial setting and a common interaction structure.

\begin{figure}[H]
\centering
\includegraphics[width=.80\linewidth]{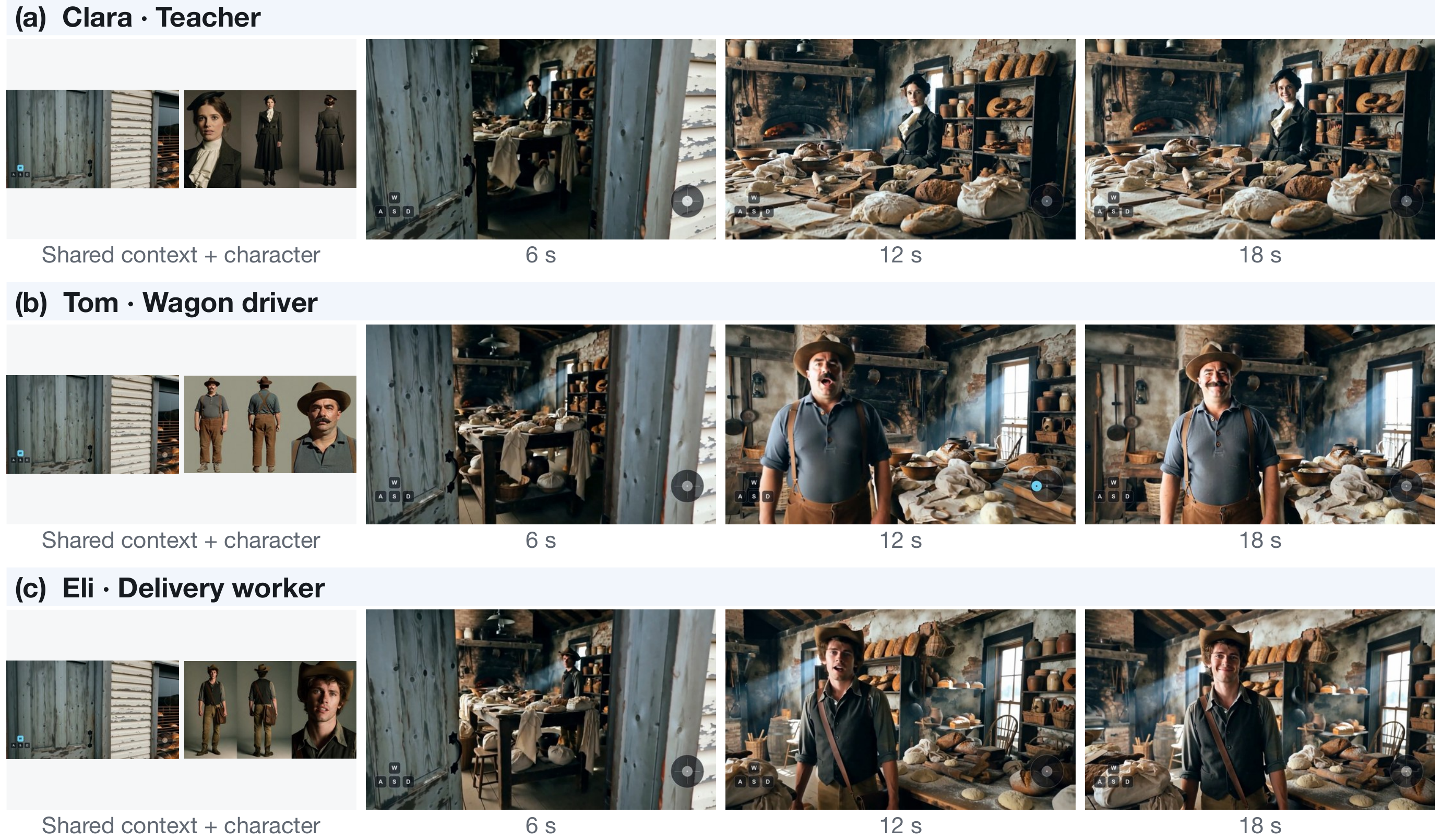}
\caption{\textbf{Reference-conditioned branching.} A shared bakery opening is paired with character references for Clara (a), Tom (b), and Eli (c), producing different encounters.}
\label{fig:perf-reference-choices}
\end{figure}

\paragraph{Embodied interaction.}
Figure~\ref{fig:perf-embodied} presents generated observations from a household robot's viewpoint. The sequences show manipulators placing a mug, retrieving a ball for a dog, and watering a plant. Successive observations connect these actions to visible changes in nearby objects within a continuing scene. These generated scenarios illustrate a potential observation interface for embodied agents; they do not measure physical robot task success.

\begin{figure}[H]
\centering
\includegraphics[width=.80\linewidth]{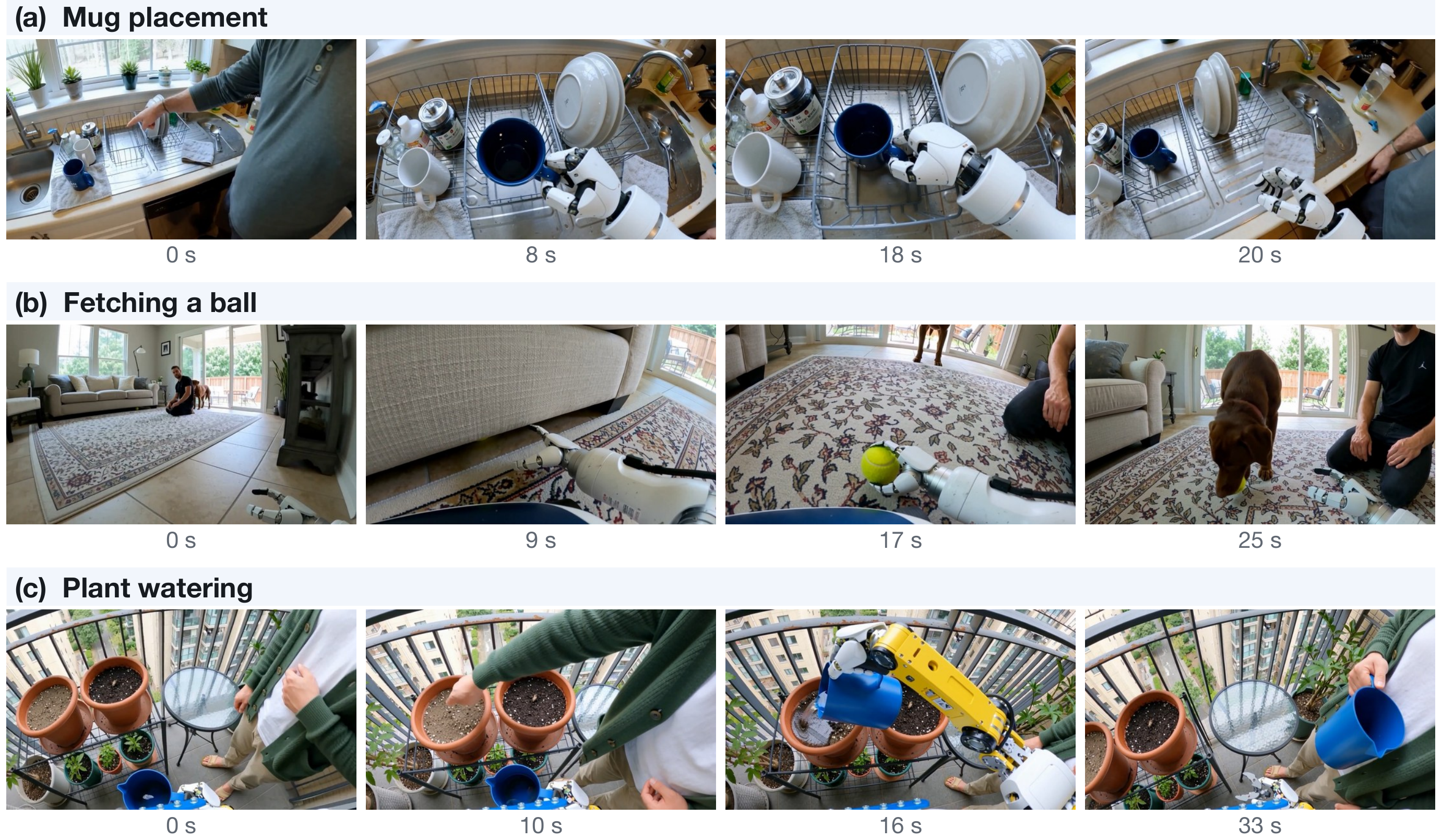}
\caption{\textbf{Embodied interaction.} Generated first-person observations of mug placement (a), ball retrieval for a dog (b), and plant watering (c), viewed from a household robot.}
\label{fig:perf-embodied}
\end{figure}

% Keep the discussion in the normal text flow and float each DAO figure separately.
\subsection{DAO-World Engine Performance}
We evaluate DAO, the world engine of XGEN, through qualitative system demonstrations and ablation experiments, focusing on autonomous agent behavior, first-person views within a shared world, and persistent world-state evolution.
\par\smallskip
\begin{figure}[!htb]
\centering
    \includegraphics[width=\linewidth]{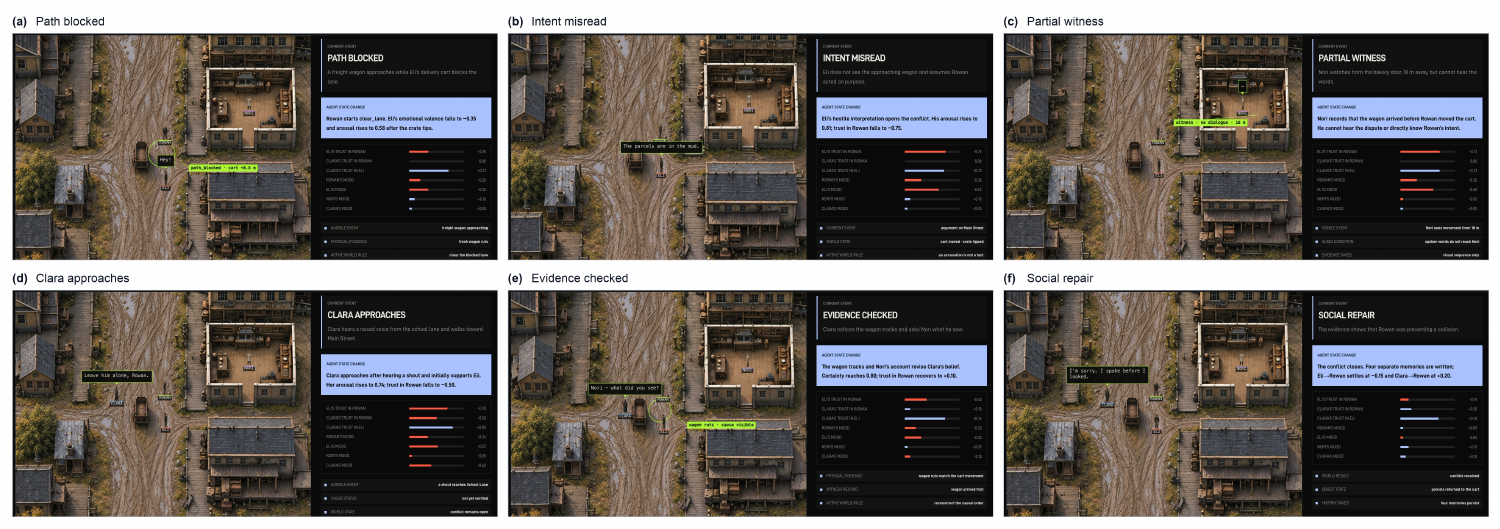}
    \caption{\textbf{Agents act autonomously and change the world.}
    Six successive events driven by agent intentions in DAO:
    (a) path blocked, (b) intent misread, (c) partial witness,
    (d) Clara approaches, (e) evidence checked, and (f) social repair.
    Local observations and accumulated evidence guide autonomous actions
    from conflict to reconciliation.}
    \label{fig:dao-autonomous-agents}
\end{figure}

\Needspace{3\baselineskip}
\paragraph{Agents autonomously.} Unlike interaction schemes that require explicit user input to trigger events, DAO supports agents that initiate actions autonomously according to the world's configuration, their own memories, local observations, and available information. Figure~\ref{fig:dao-autonomous-agents} illustrates autonomous, intention-driven interaction among DAO agents. Rowan moves a handcart to clear a blocked lane (a), but Eli interprets the displaced parcels as deliberate interference, triggering an argument (b). Nori witnesses part of the event (c), and Clara intervenes on Eli's behalf (d). Physical traces and witness evidence subsequently clarify Rowan's intention (e), leading to reconciliation (f). This progression arises from agents pursuing their own intentions and revising their actions in response to local observations and accumulated evidence, while DAO maintains the evolving shared world state.

\begin{figure}[!htb]
\centering
    \includegraphics[width=\linewidth]{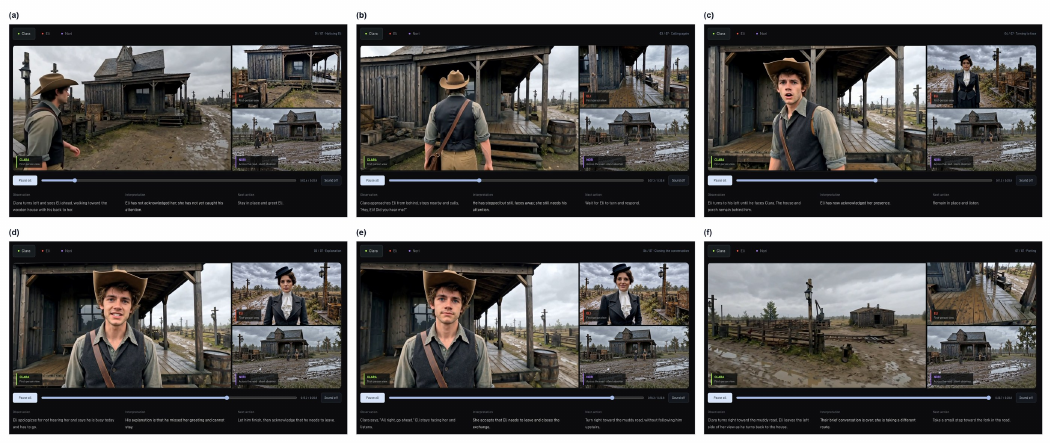}
    \caption{\textbf{Different viewpoints of agents in a shared world.}
    A shared interaction, shown from the
    viewpoints of Clara, Eli, and Nori.}
    \label{fig:dao-different-experiences}
\end{figure}

\Needspace{3\baselineskip}
\paragraph{One event, different experiences.} DAO supports first-person observations from different agents at the same simulation time, grounding their views of the world and its ongoing interactions in a shared temporal and world-state context. Figure~\ref{fig:dao-different-experiences} presents the same interaction from the first-person perspectives of Clara, Eli, and Nori. Clara and Eli experience their encounter from opposing viewpoints, while Nori observes it from across the road. DAO derives observer-specific information from a common world state and conditions JING to generate each character's local visual experience.

\begin{figure}[!htb]
\centering
    \includegraphics[width=\linewidth]{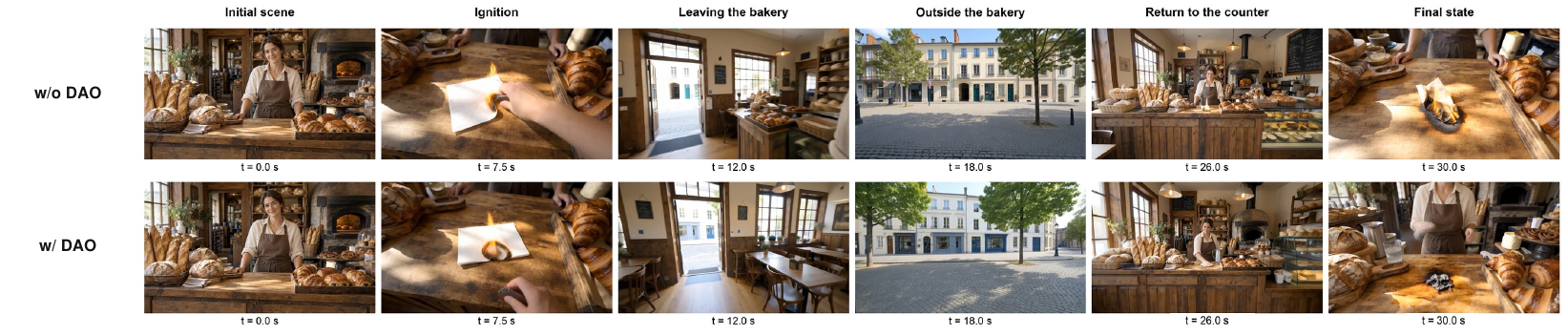}
    \caption{\textbf{JING generation results with and without DAO.}}
    \label{fig:dao-bakery-world-evolution}
\end{figure}

\Needspace{3\baselineskip}
\paragraph{Grounding JING in an evolving world.} DAO provides JING with a persistent, continuously evolving world that maintains an explicit state and supports agents with memory and autonomous, intention-driven behavior. Figure~\ref{fig:dao-bakery-world-evolution} compares JING generations with and without DAO in a departure-and-revisit scenario. Without DAO (top row), the generated revisit depicts the previously ignited napkin as still burning, with no visible evidence of an intervening agent response. With DAO (bottom row), autonomous agent--environment interactions update the world state during the observer's absence: the attendant extinguishes the fire, and JING conditions the revisit generation on the resulting state. This qualitative comparison illustrates how ongoing world simulation allows actions outside the observer's field of view to influence subsequent observations, supporting causal continuity across departures and revisits.

% Continue directly into the next section instead of forcing a mostly empty page.
\endgroup

\FloatBarrier
\Needspace{7\baselineskip}
\section{Conclusion}
\label{sec:conclusion}
We present XGEN, a Generative World Simulation framework for sustained human and agent interaction that integrates DAO, a computable world engine, with JING, an Egocentric Interactive Experience Model. DAO maintains and updates shared world state while preserving the consequences of actions; JING generates continuous first-person audiovisual experiences conditioned on references, observation history, and current actions. Together, they connect actions, state updates, and observations in a continuous loop, enabling exploration, object manipulation, and character communication grounded in prior interactions. By coupling world-state evolution with individual experiences, DAO and JING extend audiovisual generation toward simulation of the world behind it. Our long-term goal is the persistent shared world envisioned in OASIS, where encounters carry history, choices have lasting consequences, and social activity continues beyond any individual's presence. In this vision, humans and agents gain experience and shape a world that evolves through their actions and continues to develop in their absence. Future work will focus on improving motion robustness and long-term memory, reducing response latency, and strengthening consistency between shared world state and generated audiovisual content.

\begingroup
\small
\setlength{\bibsep}{5pt}
\bibliographystyle{unsrtnat}
\phantomsection
\addcontentsline{toc}{section}{References}
\bibliography{references}
\endgroup

% \appendix
% \section{Reproducibility Details}
% \label{sec:appendix}
% Use the appendix for configuration details, extended results, additional qualitative examples, and derivations that would interrupt the main narrative.
% \begin{table}[htbp]
% \centering
% \caption{\textbf{Configuration record.} Complete these fields for every reported experiment.}
% \small
% \begin{tabularx}{\textwidth}{@{}p{.31\textwidth}Y@{}}
% \toprule
% \tablehead{Item} & \tablehead{Value to record}\\
% \midrule
% Data and split & Source, version, preprocessing, and split identifier\\
% Model checkpoint & Architecture configuration and checkpoint identifier\\
% Training setup & Optimizer, schedule, batch size, and random seeds\\
% Inference setup & Precision, sampling settings, and output specification\\
% Compute environment & Accelerator, device count, and software versions\\
% Evaluation protocol & Metric implementation, sample count, and aggregation\\
% \bottomrule
% \end{tabularx}
% \end{table}

\clearpage

\section{Contributors and Acknowledgments}
All contributors are listed in alphabetical order by their last names.

\textbf{Team XGEN}
Guohao Cai, Haoxuan Che, Frank Fan, Chenran Huang, Junjia Huang, Jiayi Jiang, Yanheng Li, Leon Liu, Shaoteng Liu, Xi Liu, Feng Nie, Zhefan Rao, An Wang, Hanwen Wang, Pengxiang Yan, Jieming Yu, Qian Zha, Wentao Zheng
\end{document}